\documentclass[twoside,11pt,preprint]{article}

\usepackage{blindtext}

\usepackage{longtable}

\usepackage{booktabs}
\usepackage[load-configurations=version-1]{siunitx} 

\usepackage{graphicx}
\usepackage{xspace}
\newcommand*{\eg}{e.g.,\@\xspace}
\newcommand*{\ie}{i.e.,\@\xspace}
\usepackage{xcolor}

\newenvironment{usecase}[1]
    {\noindent\makebox[\linewidth]{\rule{0.33\linewidth}{0.5pt}}
    \begin{center}\textsc{\small\textbf{Use case} - #1}\end{center}\em\color{darkgray}\small
    }
    {\noindent\makebox[\linewidth]{\rule{0.33\linewidth}{0.5pt}}
    }

\usepackage{dmlr2e}

\usepackage{lastpage}
\dmlrheading{ }{2024}{1-\pageref{LastPage}}{2/24; Revised xx/xx}{xx/xx}{21-0000}{Egele, R; {Jacques Junior, J. C. S.}, et al.} 

\ShortHeadings{Dataset Development}{Egele, R; {Jacques Junior, J. C. S.}, et al.}
\firstpageno{1}

\begin{document}

\title{AI Competitions and Benchmarks: Dataset Development\thanks{This document is a preprint version of the 3rd Chapter of the book: \href{https://sites.google.com/chalearn.org/book/home}{Competitions and Benchmarks, the science behind the contests}.}}


\author{\name{Romain Egele}\thanks{These authors contributed equally to this work.} \email{romain.egele@universite-paris-saclay.fr}\\
       \addr University Paris-Saclay, France, and Argonne National Laboratory, USA
       \AND
       \name{{Julio C. S.} {Jacques Junior}}$^\dag$ \email{julio.silveira@ub.edu}\\
       \addr University of Barcelona and Computer Vision Center, Spain
       \AND
       \name{{Jan N.} {van Rijn}} \email{j.n.van.rijn@liacs.leidenuniv.nl}\\
       \addr Leiden Institute of Advanced Computer Science (LIACS), Leiden University, the Netherlands
       \AND
       \name{Isabelle Guyon} \email{guyon@chalearn.org} \\
       \addr University Paris-Saclay, France, ChaLearn, USA, and Google, USA
       \AND
       \name{Xavier Bar\'o} \email{xbaro@ub.edu} \\
       \addr University of Barcelona, Spain
       \AND
       \name{Albert Clap\'es} \email{aclapes@ub.edu} \\
       \addr University of Barcelona and Computer Vision Center, Spain
       \AND
       \name{Prasanna Balaprakash} \email{pbalapra@ornl.gov}\\
       \addr Oak Ridge National Laboratory, USA
       \AND
       \name{Sergio Escalera} \email{sescalera@ub.edu}\\
       \addr University of Barcelona and Computer Vision Center, Spain
       \AND
       \name{Thomas Moeslund} \email{tbm@create.aau.dk} \\
       \addr Aalborg University, Denmark
       \AND
       \name{Jun Wan} \email{jun.wan@ia.ac.cn}\\
       \addr MAIS, Institute of Automation, Chinese Academy of Sciences, China
       }

\editor{Isabelle Guyon, Adrien Pavao and Evelyne Viegas}

\maketitle

\begin{abstract}
Machine learning is now used in many applications thanks to its ability to predict, generate, or discover patterns from large quantities of data. 
However, the process of collecting and transforming data for practical use is intricate. 
Even in today's digital era, where substantial data is generated daily, it is uncommon for it to be readily usable; 
most often, it necessitates meticulous manual data preparation. 
The haste in developing new models can frequently result in various shortcomings, potentially posing risks when deployed in real-world scenarios (\eg social discrimination, critical failures), leading to the failure or substantial escalation of costs in AI-based projects. 
This chapter provides a comprehensive overview of established methodological tools, enriched by our practical experience, in the development of datasets for machine learning. 
Initially, we develop the tasks involved in dataset development and offer insights into their effective management (including requirements, design, implementation, evaluation, distribution, and maintenance). 
Then, we provide more details about the implementation process which includes data collection, transformation, and quality evaluation. 
Finally, we address practical considerations regarding dataset distribution and maintenance.
\end{abstract}

\section{Introduction}

In today's digital world, large amounts of data are generated every day in a wide range of domains.
Using this data, machine learning methods can be used to train AI models that address or automate various tasks.
As machine learning is used widely both in research and industry, following the wrong procedures in collecting and processing a dataset can lead to various downstream problems when models are being trained on this data, \eg problems with privacy or fairness. 
In this chapter, we present a framework that aims to help in developing a dataset in a more principled way as well as identify core actions to be performed for better management of such a project.

\begin{figure}[htbp]
    \centering
    \includegraphics[width=1.0\textwidth]{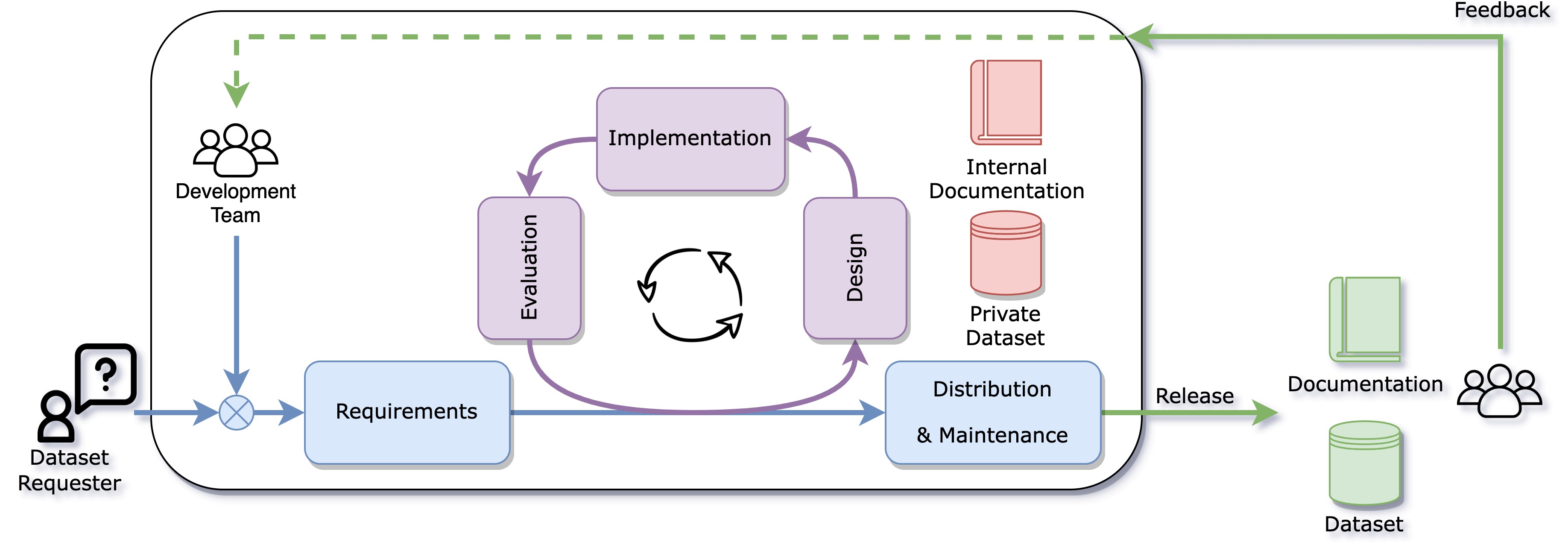}
    \caption{The dataset development cycle.
    \label{fig:dataset_development_lifecycle}} 
\end{figure}

As mentioned by \citet{10.1145/3442188.3445918}, dataset development is not a linear process that has all detailed specifications from the start. It can be structured using an agile\footnote{Agile is a common term from software development.}~\citep{chin2004agile} management methodology with core components interacting with each other as well as evolving iteratively. A representative cycle of dataset development is presented in Figure~\ref{fig:dataset_development_lifecycle}. One cycle is composed of five components: the {requirements} analysis involves the principal stakeholders and consists in defining the needs of the developed dataset; the {design} involves the domain expert and consists in defining how to structure the dataset and its implementation; the {implementation} involves data creators (\eg data/software engineers, labelers) and consists in collecting and transforming the data to be usable; the {evaluation} involves data scientists and adversarial testers, consists in assessing the quality of the developed dataset with respect to its requirements; the {distribution} and {maintenance} involve regulation, storage, and network experts and consist in defining the storage and accessibility of the dataset.

During this cycle, different aspects need to be documented. The dataset development team should 
document as part of an internal (in the sense that it is meant to be used by the development team) documentation of common assumptions made when developing the dataset (\eg explaining why the collected data are representative of the target population) as well as the tools or processes developed to acquire the data (\eg survey, software). 
In addition, the development team should maintain public documentation explaining the content of the dataset, its purpose, and its technical usage (\eg software to install, interface to retrieve data samples, or meta-data of features). 
This public documentation differs from internal documentation in the sense that it could contain fewer details, and sensitive information (such as personal identifiers) should be left out.
Typically, both private (raw) versions of the data, as well as public releases of the data exist.  

The framework we present aims to support the development of datasets used in machine learning. 
While the dataset development cycle consists of many stakeholders and roles, in smaller dataset development projects a small group of persons can cover multiple roles. 
We substantiated our claims with scientific literature when possible. 
However, some paragraphs could not be matched with such references but are the results of our personal experience (\eg data challenges).

\section{Documentation}
\label{chap3:sec:documentation}

We first review various standards of building documentation (either public or private) of the development process and the final produced dataset. Documentation is important and spans every aspect of the dataset development in order to have better transparency and traceability which will improve trust and safety. Documentation must be performed with reflexivity which means its goal is to clarify or uncover obscure discretionary decisions (\eg power dynamics, differences of perception) in order to understand their effect on the produced dataset~\citep{Miceli_FAccT21}. In an effort to systematize the documentation of datasets, different initiatives emerged such as new methods~\citep{10.1145/3458723,Bender:2018,Holland:2018}, best practices\footnote{Meta-data and data documentation: \href{https://tinyurl.com/5j5ynu6p}{tinyurl.com/5j5ynu6p}}, and formats\footnote{AutoML format: \href{https://tinyurl.com/yc5uk4xh}{tinyurl.com/yc5uk4xh}}. The main aspects addressed in dataset documentation include its purpose, composition, collection process, preprocessing, intended usage, distribution, and maintenance~\citep{wilkinson_fair_2016}. 

Several standards for dataset documentation have been proposed. 
\begin{description}

\item[Datasheets for datasets:] \citet{10.1145/3458723} proposes specific example problems to document around the core components of the dataset development. It encourages every dataset to be accompanied by documentation of its motivation (requirements), composition (design), collection process (implementation), recommended uses, distribution, and maintenance. These guidelines are adopted by the NeurIPS Dataset and Benchmark track. Similarly the DC-Check (DC standing for: data-centric) framework was released~\citep{dccheck_seedat_20222} to spawn a broader range of data-centric related tasks on a general range of applications. 
\citet{Bender:2018} propose similar guidelines called ``data statements'', with a focus on natural language processing applications.

\item[FAIR principles:] \citet{wilkinson_fair_2016} focus on distribution and maintenance aspects and provide guidelines to improve the findability, accessibility, interoperability, and reuse of datasets. This initiative aims to standardize digital practices around dataset distribution and maintenance to improve the automation of software using data and favor the reuse of datasets.

\item[Dataset nutrition label:] \citet{Holland:2018} follow the idea of nutrition facts labels in order to help create a summarized diagnosis of the quality of a dataset. It comprises diverse qualitative and quantitative modules generated through checklists or multiple statistical analyses of the dataset and displayed in a standard fashion.
\end{description}

However, despite the good intentions of standard documentation practices, they often must be refined for the specificities of the target use and some attributes may not be allowed to be disclosed or stored due to data regulation (\eg race, belief) which therefore prohibits public verification of some statistical properties.

\begin{usecase}{Facial Emotion Recognition}
To illustrate documentation, we give an example of potential aspects to be documented when developing a video dataset for facial emotion recognition~\citep{7374704}. In this context, videotapes are recorded from a group of participants.
\begin{itemize}
    \item What is the dataset intended to be used for (\eg for which application and for which population)? -- \textbf{Requirements}
    \item How are participants recruited for video recording (\eg gender, age, hairstyle, use of glasses)? -- \textbf{Design}
    \item How is the privacy of recorded participants managed? -- \textbf{Design}
    \item For the recording protocol: Are participants following a script during recordings? Are participants stimulated by a particular incentive during recording? -- \textbf{Design} 
    \item For the annotation protocol: How are the annotation defined (\eg categorical, discrete, or real values)? How are annotators selected (\eg gender, age, etc.)? How are data to be annotated defined (\eg per frame, per video segment, with or without sound)? How many annotators observe each data? -- \textbf{Design}
    \item How are acquired data transformed (\eg calibration)? How are final labels computed? What is the definition of frames and sample rate? -- \textbf{Implementation}
    \item For the investigation of social bias you may document some sensitive information regarding participants and annotators (\eg gender, age, race) while being cautious to respect data regulation, privacy, and consent. What are the possible biases from the selected populations (participants, annotators)? How is the annotators' agreement evaluated (\eg metric) and what is its value? -- \textbf{Evaluation}
    \item Where are the data hosted? How can the data be accessed? -- \textbf{Distribution} \& \textbf{Maintenance}
\end{itemize}
\end{usecase}

Finally, we give some practical advantages of good dataset documentation. The {dataset development team} can benefit from the following aspects: a better management of the dataset development by an improved understanding of why, how, and what is done to produce the dataset; a better traceability of possible bugs or flaws; an improved reusability of developed tools due to clear documentation and principled development; a reduced presence of flaws in the produced dataset; a better dataset quality. On the other hand, the {dataset users} can benefit from the following aspects: an improved usability of the dataset; an improved understanding and trust of the dataset; a better quality of machine learning models; and a better reporting of possible flaws discovered in the data.

Nevertheless, while documentation helps to improve dataset quality and therefore the quality of machine learning models using them, some limitations still exist. Companies often regard some of the information that could or should be documented as confidential, especially if it involves details about the intended product or if some of the processes involved in producing the dataset are a strategic advantage~\citep{Miceli_FAccT21}. For this reason, we differentiate private (required for audits) and public documentation (required for the user). Then, documentation is often seen as time-consuming work that is likely to delay the completion of other tasks that are perceived as more important. It is often perceived as an optional, nice-to-have but not must-have, component that is implemented last~\citep{Miceli_FAccT21}. 
Lastly, producing complete but synthetic and clear documentation is challenging. The format of documentation may vary for the different stakeholders (engineers, statisticians, business analysts) and create redundancy (pdf document, book, website). We encourage dataset development teams to use tools such as Sphinx\footnote{\href{https://www.sphinx-doc.org/en/master/}{www.sphinx-doc.org}}, ReadTheDocs\footnote{\href{https://readthedocs.org/}{readthedocs.org}} and Pandoc\footnote{\href{https://pandoc.org/}{pandoc.org}} to automate the build of documentation and navigate between formats.

\section{Requirements}
\label{chap3:sec:requirementsanalysis}
The requirements analysis is the first step of the dataset development cycle (see Figure~\ref{fig:dataset_development_lifecycle}) where the dataset requester (representing the main stakeholder requiring the dataset) and the dataset development team (representing who is in charge of producing the dataset) meet to define the requirements. During this phase, the following topics can be addressed: 
\begin{enumerate}
    \item Why is the dataset needed?
        \begin{itemize}
            \item \textbf{Application scenario:} What are the intended purposes and use cases?
            \item \textbf{Machine learning tasks:} What type of machine learning techniques (\eg supervised, unsupervised, reinforcement learning) is planned to be used? How will tasks be carved out of data? 
            \item \textbf{Users:} Which group of users do we expect to use the dataset?
        \end{itemize}
    \item How is the dataset developed?
        \begin{itemize}
            \item \textbf{Prior work:} Are there already existing datasets filling this need (see Section~\ref{chap3:sec:reuse} and \ref{chap3:sec:gathering} about potential sources of already collected data)? Will collecting (more) data solve the problem at hand, or help to have a better understanding of it?
            \item \textbf{Method:} What dataset development protocol is planned to be used? 
            \item \textbf{Ethics:} Are the intended purposes ethical? What are possible fairness and privacy issues and how are they going to be evaluated?  Is collecting such data considered experimenting on human subjects? 
            \item \textbf{Risks:} What adverse usage could be done from the dataset? Are the risks worth the trouble? How can the risks be reduced?
            \item \textbf{Constraints:} What are the anticipated difficulties limiting the development of the dataset (\eg recruitment of subjects)? In the case of data involving human subjects or a legal entity, it is crucial to follow regulations from governmental regulations such as GDPR~\citep{10.5555/3152676}. Sometimes, the data needs to be anonymized, cannot be stored anywhere (risk of data leaks), and its legal framework (\eg application, lifespan) needs to be defined before acquisition.
            \item \textbf{Development team:} Who is composing the dataset development team? Who will lead the effort? How are the roles distributed (design, implementation, evaluation, distribution, and maintenance)?
            \item \textbf{Resources:} How many resources will be required to complete the whole ddlc, including compensation of staff, recruitment of volunteers, payment of annotation services, computational resources, etc.? Also, consider the environmental impact (CO2 emissions).
        \end{itemize}
    \item What is the dataset expected to be?
        \begin{itemize}
            \item \textbf{Content:} What information does the dataset needs to contain (\eg features, annotations, meta-data)?
            \item \textbf{Baseline:} What baseline modeling methods will be used to evaluate the quality of the dataset? What metrics of success will be used, including utility, fairness, and privacy?
            \item \textbf{Ownership:} Who will own the rights? Who will be legally responsible?
            \item \textbf{Storage:} Where does the dataset need to be hosted?
            \item \textbf{Distribution:} How is the dataset going to be accessed (\eg through a web API)?
        \end{itemize} 
    \item When is the dataset expected to be delivered?
\end{enumerate}

In fact, a dataset cannot be designed without considering its envisioned purpose. Also, the interplay between the dataset and the machine learning method used is also essential. A dataset represents a snapshot of the real world used to train (and possibly evaluate) a learning algorithm on a particular task. 
It is essential for quality assurance to involve baseline modeling methods and their performance evaluation early in the dataset development process. If the utility, fairness, ethical concerns and privacy of the data cannot be assessed in the context of an actual learning task, the dataset will likely end up being useless.

Regarding ethical considerations, it is advisable to create a committee including diverse members in terms of competence, demographics, and cultural backgrounds. The committee should include persons competent in the target application area, machine learning/data science, and persons representative of the subject and/or target population (if human subjects are involved in data collection or are affected by data usage). Additionally, it is advisable to include an ethics expert and a law expert. Ensuring that one member at least is not affiliated with the organization creating the dataset and that no member has a conflict of interest. In the US, such committees are called Institutional Review Board\footnote{Institutional Review Board: \href{https://tinyurl.com/mvpd292w}{https://tinyurl.com/mvpd292w}} and can be registered officially. Also, some general guidelines\footnote{Ethics Guidelines for Trustworthy AI: \href{https://tinyurl.com/2zc7hfdz}{https://tinyurl.com/2zc7hfdz}}$^,$\footnote{Recommendation on the Ethics of Artificial Intelligence: \href{https://tinyurl.com/3ew8xp4e}{https://tinyurl.com/3ew8xp4e}} can be followed.

\section{Design}
\label{chap3:sec:design}

The design phase is about defining more precisely what the dataset should contain, and how it will be implemented (\ie collected and transformed), evaluated, distributed, and maintained. The complexity of implementation design, which comprises data collection and transformation, is directly related to the choice of creating a dataset from scratch or reusing, repurposing, and recycling existing data. For many use cases, processes describing dataset development already exist and are published such as at the NeurIPS Dataset and Benchmark track\footnote{NeurIPS Datasets and Benchmarks track: \href{https://tinyurl.com/37u4cbx3}{https://tinyurl.com/37u4cbx3}}. We recommend exploring this literature to look for dataset development specific methods.

The design of a dataset is often tedious while being key to innovation. Of course, designing a dataset directly results from the definition of requirements but, it is also possible to update the requirements based on findings of the design step. In fact, any critical aspect missed at the requirements and design stages may compromise the whole dataset development process. Even though these steps should be designed thoroughly it is likely that flaws are discovered later. For instance, in the case where data are planned to be collected through an online survey. The development team can forget to ask participants to sign an agreement, which gives the consent to process, use the data, transfer rights including copyright, and as such collected data cannot be used. Therefore, we encourage the development team to bootstrap quickly the whole development cycle in order to help refine requirements and design. As a result, the dataset development should not be a linear closed loop but an iterative and interactive process.

When developing a dataset, either observational (\ie the variables of the phenomenon cannot be controlled) or experimental (\ie the variables of the phenomenon can be controlled) data can be used. Similarly, data can be collected {de novo} (\ie from scratch) or from existing sources (\ie reuse, repurpose, and recycle).

\subsection{De Novo Data}
\label{chap3:sec:denovo}

In this section, we describe design aspects for {de novo} data. The first setting to consider is which variables of the target phenomenon can be controlled or not and in which quantity they appear (\eg difference between pixels of images and tabular dataset).
We distinguish between an experimental case (where variables can be controlled) and an observational case (where variables can not be controlled). 
In the experimental case, when only a few variables are available, one can decide to discretize them and explore all possible combinations.
When larger quantities of variables are available, one often resorts to random sampling. 
In the observational case, even though variables cannot be controlled and impacting factors are often intractable (\eg describing a patient in healthcare) random sampling is also advocated (random trials) in order to vary these factors. In this case, it is important to check that the observed population is indeed randomly sampled and not biased (\eg selection bias). Other considerations for the dataset design are:

\begin{description}

    \item[Data quantity:] How many samples should the dataset have? An example of character recognition is proposed in~\citep{Guyon96whatsize}. However, it may be difficult to have such information beforehand. It is possible to refine the quantity needed by involving the modeling process during development (\ie baseline in the evaluation step). A default choice is to collect as much data as possible and leave the choice of quantity to the user.
    
    \item[Data balancing:] How to make sure that you have enough samples of each group that should be represented? For instance, depending on the application, you may want to balance groups by gender, age, or educational background. It is advised to pay attention to the possible need to consider cross-sectional groups' gender $\times$ age $\times$ educational background. Unfortunately, the larger the number of grouping factors to be considered, the larger the number of samples to be collected to adhere to a minimal group size.
    
    \item[Data annotations:] Are labels required? Do we need more advanced annotations, such as bounding boxes around the subject? If yes, how is the labeling process operated? For example, this can be achieved through crowd-sourcing, citizen science, or commercial parties. When making use of such services, one should always carefully check the conditions under which these labeling operations are being performed (\eg are the persons performing the labeling doing this under fair work conditions), and whether this process can be (semi-)automated.
    
    \item[Data representation:] How are data represented? Many data structures exist to represent data, it can be tables, images, videos, text files, and/or graphs. The data can be compressed or not. The data can be accessed incrementally or all at once (we refer the reader to the HDF5 open format and library\footnote{\href{https://www.hdfgroup.org/solutions/hdf5}{https://www.hdfgroup.org/solutions/hdf5}}). Can we provide a data reader? If tabular data are collected, what is the set of features to collect?
    
    \item[Meta-data:] What meta-data can be collected (\eg date of recording, operator name, temperature)? Generally, the more meta-data, the better. Meta-data can help identify bias or spurious correlations. The meta-data explain the context of generated data and therefore it can help determine if the data was well collected in diverse settings (\eg at different times or temperatures). Also, meta-data should not be correlated with the predicted variable. If this is the case, then some spurious correlation can be identified and resolved by modification of the dataset creation process.
\end{description}

\subsection{Reusing, Repurposing, and Recyling Data}
\label{chap3:sec:reuse}

As discussed before, {de novo} dataset development can be a tedious process. However, many datasets are now publicly accessible~\citep{koch2021reduced} with a license allowing to use them free of charge (\eg Creative Common\footnote{\href{https://creativecommons.org}{https://creativecommons.org}}). In addition, search engines (a short list is available at the end of this section) can help find datasets corresponding to specific needs. Therefore, before performing {de novo} data collection, it is important to investigate such opportunities which can help save a considerable amount of resources. We refer to such methods as {data reusing} (analogously to reusing a plastic bottle of water by refilling it), {data repurposing} (analogously to repurposing a plastic bottle of water to collect leaking water from a pipe) and {data recycling} (analogously to breaking down the plastic bottle of water to make plastic boxes). These concepts are ordered from fewer to more modifications applied to the pre-existing data, however, the boundaries among them are not clear and can overlap.

On the one hand, data reusing is the practice of directly reusing data (\ie without modification, including its purpose) when the use case is similar and the required information is already available. Hence, we talk about data reusing when the initial product (\ie the data) does not need to be transformed before being ingested and the intended purpose remains the same.

Then, one may re-use the exact same set of data but re-purposing it from being used in research to commercial applications. In this case, many ethical and privacy concerns need to be revisited, among other aspects. 

When leveraging existing data it is important to be careful with possible {deprecation}, {bias}, and {retirement} of the source data. A dataset becomes deprecated when it is still publicly accessible, but for some reason (\eg social bias) it should not be used in practice. An example of a deprecated dataset is the Boston house prices dataset
~\citep{harrison1978hedonic}, which is kept public for scientific traceability and reproducibility as well as social bias study but discouraged from being used otherwise. In this case, the deprecation was due to the presence of social bias in the data. Similarly, some datasets can be retired (\ie removed from public access) such as the Tiny Images dataset\footnote{Tiny Images dataset: \href{https://tinyurl.com/2vfa4xve}{https://tinyurl.com/2vfa4xve}}~\citep{torralba200880} due to the presence of derogatory terms as categories and offensive images. 

In some cases, it is required to perform light data transformations in order to enable data repurposing. For example, if the to-be-predicted target variable changes, a re-annotation process may be performed. In other cases, pre-existing data can be completed by new samples and/or features. For instance, the ``First Impressions V2''~\citep{Escalante:IJCNN:2017} dataset is a typical example of repurposing. The original ``First Impressions''~\citep{lopez2016chalearn} dataset was annotated with Big-Five personality traits with the purpose of analyzing personality from audio-visual data. In the ``V2'' version the audio-visual data was kept the same, but new labels were included, like an additional ``interview'' variable and transcriptions, with the purpose of advancing the state of the art on explainable machine learning. 

Last but not least, requiring additional effort, data recycling leverages existing material with the possibility of reshaping it entirely. In this case, the dataset's purpose can further differ from its initial purpose. Therefore, it has to be verified that it is allowed by the license of the dataset. 
This is why recycling requires extra work while remaining less tedious than {de novo} collection. An example of recycling is the creation of an image dataset with a single face per image per sample given a pre-existing image dataset containing single or multiple people on each image with full or partial bodies~\citep{agustsson2017appareal}.

Examples of {Datasets Search Engines and Providers} are \href{https://datasetsearch.research.google.com}{Dataset Search from Google}, \href{https://www.kaggle.com}{Kaggle}, \href{https://www.openml.org}{OpenML.org}~\citep{vanschoren2014openml,bischl2021openml}, \href{https://archive.ics.uci.edu}{UCI Machine Learning Repository}, \href{https://huggingface.co/datasets}{Hugging Face Datasets}, and the \href{https://nips.cc/Conferences/2021/DatasetsBenchmarks/AcceptedPapers}{NeurIPS: Datasets and Benchmarks} track.

\section{Implementation}
\label{chap3:sec:implementation}

\begin{figure}[!tb]
    \centering
    \includegraphics[width=0.75\linewidth]{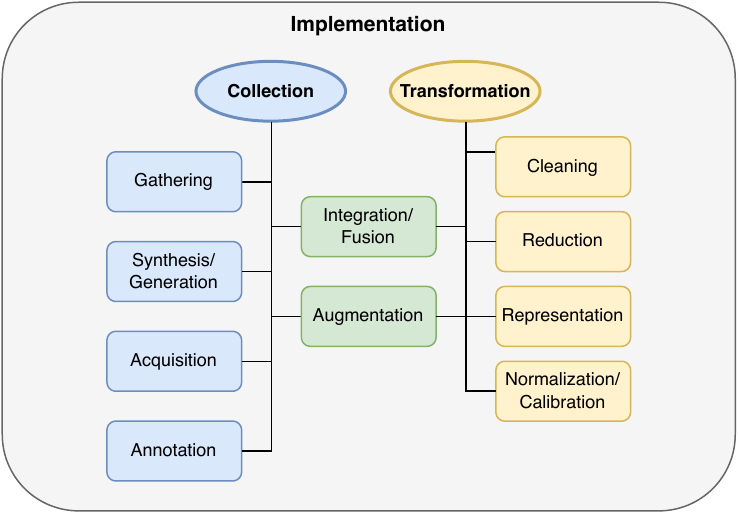}
    \caption{Categorization of sub-tasks included in data collection and transformation. 
    Collection operators (blue) take as input a design and as output a dataset (of an arbitrary size). 
    Transformation operators (yellow) require as input a dataset and will have as output also a dataset. 
    Some operators (green) fall into both categories. 
    A typical data development process combines several of these operators into a pipeline, always starting with a collection operator.}
    \label{fig:dataset-development-implementation-taxonomy}
\end{figure}

This section focuses on the implementation of the dataset development. 
It is a set of processes, which are described in Figure~\ref{fig:dataset-development-implementation-taxonomy}.
This categorizes possible sub-processes into two categories: 
Collection processes (blue) take as input a design (Section~\ref{chap3:sec:design}), and as output a dataset (of an arbitrary size). 
Transformation processes (yellow) require as input a dataset and will have as output also a dataset. 
Some processes (green) fall into both categories. 

These processes are the functions that will produce a dataset on which machine learning models are trained. The implementation is divided into sub-tasks which we group around two axes: data collection and data transformation. Data collection entails the {gathering}, {acquisition}, {synthetic data generation}, and {annotation} of data. To avoid any confusion with other works that interchangeably use the words ``collection'', ``gathering'' and ``acquisition'' to denote the same thing, this chapter will adopt the word {gathering} for the cases where the data already exist {digitally}, and the data creator is obtaining it from somewhere (\eg collecting images from Google images or text from Wikipedia), and the word {acquisition} for the cases where the data is captured through any type of sensor (\eg recording someone with a camera, as in the case of facial emotion recognition, any type of information obtained from questionnaires/surveys, or digitization of documents). Data transformation includes {cleaning}, {reduction}, {representation}, and {normalization} or {calibration} of data.
The tasks of data {integration} or {fusion}, and {augmentation} can be categorized as both collection and transformation operators; we therefore link them to both in Figure~\ref{fig:dataset-development-implementation-taxonomy}.

\begin{figure}[tb!]
    \centering
    \includegraphics[width=0.65\linewidth]{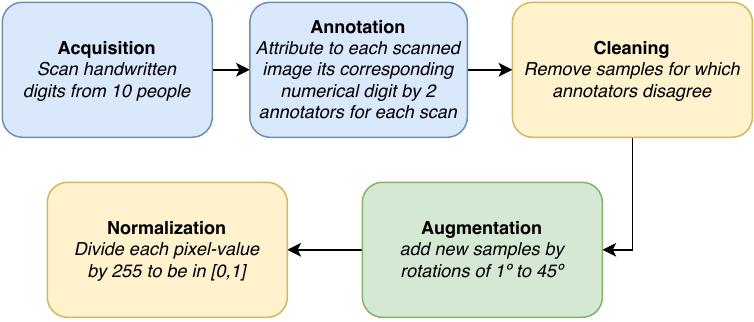}
    \caption{An example flow chart diagram of a data implementation pipeline that creates a dataset for handwritten digits classification.}
    \label{fig:example-flow-chart-data-implementation}
\end{figure}

The implementation phase typically contains various sub-processes, defined by the design.
An example of such a pipeline of processes is visualized in Figure~\ref{fig:example-flow-chart-data-implementation}, in which a pipeline combines 5 sub-processes. 

Note that data evaluation can also be considered in the case of a dynamic data collection process, for example, to keep collecting/augmenting data until a pre-defined objective is satisfied. Section~\ref{chap3:sec:evaluation} details on these evaluation processes. 
Next, we discuss each part (collection and transformation) in more detail.

\subsection{Data Collection} \label{chap3:sec:datacollection}
Data collection is the set of processes, which can gather, generate or measure information in order to create a dataset. Therefore, data collection includes data gathering, data synthesis, data acquisition, and data annotation. 

Any of these processes can be performed in an observational or experimental setting. 
In the observational setting, the investigator in charge of data collection does not interfere with the phenomenon. The distribution of collected samples is supposed to reflect the natural distribution of data. For example, a naturalist studying wildlife may decide to set up a camera trap in a forest to take pictures of animals living in that area.
In contrast, in the experimental setting, the investigator may interfere with the phenomenon to achieve desired effects. The distribution of samples collected follows an experimental plan or design. For instance, a pet food company may want to study the influence of certain dog foods on certain dog breeds and conduct a trial, assigning different regimens to dogs of various breeds, and then evaluating their energy by videotaping them. 
There also exists in-between cases in which data are observational, but the investigator samples data and/or features in an active way~\citep{settles2009active}. For example, a photographer going on a photo safari may use aesthetic criteria to make their shots.

When performing data collection it is often the case that meta-data are available or can be created to describe processed data. It is important to save as much meta-data as possible to perform better data evaluation later.
The question of how much data should be collected is often addressed by using learning curves of machine learning algorithms~\citep{mohr2022learning,mohr2023fast}.

\subsubsection{Gathering} \label{chap3:sec:gathering}

Data gathering is the set of processes by which data are brought together from sources where the data is already stored digitally and the acquisition (Section~\ref{chap3:sec:dataacquisition}) process cannot be influenced (see, \eg \citet{ullah2022metaalbum}). 
Through the emergence of the modern web where one can quickly access a massive amount of public data, the cost can be relatively low. 
These techniques suffer from high noise (\eg picture wrongly tagged and therefore wrongly returned by a search engine) as well as ethical and regulatory limitations (\eg privacy, copyrights, license).

The first way of performing data gathering is through the use of a search engine (\eg Google Image, Bing) in which case it is important to comply with the regulations.
Another way is through web scraping, which is the practice of automating the search and downloading of data from the Internet. It provides more granularity to develop the gathering process. Tools such as {Scrapy}\footnote{\href{https://scrapy.org}{https://scrapy.org}} can help to set up such a process. However, it must be performed responsibly such as by following robot policies\footnote{How to write and submit a robots.txt file: \href{http://tinyurl.com/a9tvx6n8}{http://tinyurl.com/a9tvx6n8}}, which in some cases restrict access to robots. 

Last but not least, the gathering can be performed through crowd-sourcing~\citep{Roh:2021,Garcia-Molina:2018} where human workers are generally given micro tasks to gather bits of data that collectively become the generated dataset. More generally, it can be defined as the process of obtaining needed services, ideas, or content by soliciting contributions from a large group of people~\citep{Zhang:2016}.
While crowd-sourcing can be applied both to data gathering and annotation, the methodologies applied differ, and its application to data annotation is detailed in Section~\ref{chap3:sec:dataannotation}. Crowd-sourcing to gather data can be performed implicitly or explicitly~\citep{Garcia-Molina:2018}. It is implicit when people are not aware, such as through website analytics. For example, by watching a movie an individual provides data about the popularity of this movie. Then, the gathering is explicit when subjects get a request for information. The machine learning community has been using crowd-sourcing as a tool to gather massive amounts of data, which can be considered a prominent way to outsource work to people reachable online. One of the most popular platforms for crowd-sourcing in machine learning at the time of writing is Amazon Mechanical Turk\footnote{\href{https://www.mturk.com}{https://www.mturk.com}}, where tasks are assigned to remote workers, which are compensated when the task is completed~\citep{Roh:2021}.
Of course, it is also the responsibility of the dataset collector to check that the workers who perform this labeling are doing so under ethical work conditions. 

\subsubsection{Synthesis and Generation} 

Data synthesis, also known as synthetic data generation, is the process of generating artificial data that mimics real-world observations and can be used to (pre)train machine learning models when actual data is difficult or expensive to get. It has become an attractive field of study because of data-hungry technologies such as deep learning~\citep{Nikolenko:2021}. Synthetic data can be generated procedurally~\citep{Queiroz:2010,wood2021fake} (\ie predefined set of rules), through simulations~\citep{Dosovitskiy17} or using generative models (\eg generative adversarial networks~\citep{Karras_2019_CVPR}).
One of the main motivations behind its use is the reduced cost to create larger datasets where factors of variability (\ie parameters of the synthetic generator) can be manipulated on demand. In addition, the generated data can often be directly annotated by leveraging the parameters of the data generator or using post-processing techniques. Other advantages include the possibility of better-controlling privacy \citep{yale2020generation,Kuppa:2021} and fairness \citep{bhanot2021problem,Bhanot:2022}.
However, one of the main limitations of synthetic data is that simulated phenomena often need to closely represent the real-world scenario. Another challenge is related to transfer learning \citep{weiss2016survey} and domain adaptation~\citep{Ajith:2023}, since a predictor trained on synthetic data (source domain) must generalize well to real data (target domain). Hence, the data used to eventually train the predictor implicitly should have similarities with the target data on which the trained model is deployed. Oftentimes, artificial data (available in large quantities) are used to pre-train the model which is later fine-tuned on real data (available in smaller quantities) before deployment. 

The level of abstraction and realism can vary depending on the application domain, context, and needs. For example, in the case of synthetic data generated through simulations and computer graphics applied to autonomous driving cars~\citep{Dosovitskiy17}, the level of abstraction and realism can be associated with the rendering quality, but also with respect to the behavior of the different simulated agents and phenomena (\eg interaction between agents, traffic, weather, etc.). Each of them has a particular impact on the outcomes if the data are used for training a machine learning method. The associated costs and resources (\eg data experts, designers, software engineers, etc.) required to achieve the desired level of abstraction and the trustworthiness of the generated data are additional barriers that might limit wider usability to train and evaluate machine learning models.

Synthetic data is also reported in the literature as a way to perform data augmentation, where both synthetic and real data are combined. However, a recent study on face analysis proposed to train their models with synthetic data only~\citep{wood2021fake}, opening up new approaches to better address fairness and privacy. 

\subsubsection{Acquisition} \label{chap3:sec:dataacquisition}

Data acquisition is the process that converts a real-world signal into a digital representation~\citep{emilio2013data}. A basic example is the acquisition of temperature data through a thermometer. Some techniques such as quantization~\citep{gray1998quantization}, signal sampling~\citep{higgins1996sampling} or real number encoding~\footnote{\href{https://ieeexplore.ieee.org/document/8766229}{https://ieeexplore.ieee.org/document/8766229}}, which have already been studied in depth in the information theory literature for physical signals, are often used for this purpose. 

Performing data acquisition usually requires a well-designed experimental protocol (Section~\ref{chap3:sec:design}). In addition, it is always important to record additional meta-data. Some basic meta-data are the measurement time, location, and model of the device used for acquisition. Meta-data can be descriptive such as an overall description of the dataset and variables (\eg units of physical quantities, and preferably following international standards). 
They can also contain copyrights or license terms. In general, meta-data is to be understood as data used to describe the dataset in order to maintain tractability about how the data were acquired. In many cases, they can become the dataset of another dataset, for example, in the case of integration/fusion. In fact, meta-data play an important role in identifying bias. 

\subsubsection{Annotation} \label{chap3:sec:dataannotation}

Data annotation is the process of mapping existing data samples to other data. It is often performed in the context of supervised learning, which includes two principal variants: classification and regression. In classification tasks, the goal is to train a model that returns one of many possible classes for each sample. In this context, data annotation is referred to as data labeling. In regression tasks, the goal is to train a model that predicts a real number given a sample. Therefore, it is much harder to annotate with continuous variables instead of a set of fixed classes~\citep{Roh:2021}, which can explain why data annotation research has mostly been focused on data labeling for classification. Other types of supervised tasks exist such as:
(i) describing the title of images (image captioning), (ii) labeling the style of a music record (classification), (iii) predicting the age of an individual (regression), (iv) rating an Amazon product by a score (categorical regression), or (v) drawing a bounding-box on an object in an image (object detection).

\citet{Roh:2021} propose the following categories for understanding the data labeling landscape: crowd-based labeling (\eg via crowd-sourcing or active learning~\citep{Tang:DSAA:2021}), automatically labeling from existing labels (\eg through semi-supervised learning~\citep{Engelen:2020}), and the use of weak labels~\citep{ratner2017snorkel} (\ie generating imperfect labels, but in large quantities to compensate for the lower quality).

First, {crowd-sourcing} techniques~\citep{Zhang:2016} are in general focused on running tasks with many workers who are not necessarily experts (either on labeling or on any particular task). Therefore, different solutions have been proposed to collect more accurate and trusted labels (\eg see~\citep{Nowak:2010,8462493}). In this line, a general procedure for controlling and ensuring the quality of data labeling is to have multiple workers annotate the same sample so that an agreement level can be computed. This way, any bias the workers may have can be identified and mitigated, with the cost of increasing time and resources. However, it does not necessarily include human perception bias, which are much more difficult to identify and mitigate (discussed in Section~\ref{chap3:sec:fairness}). Although various inter-annotator agreement measures exist~\citep{Checco2017} for simple categorical and ordinal labeling tasks, relatively little work has considered more complex labeling tasks, such as structured, multi-object, and free-text annotations~\citep{Braylan2022}. 
Providing effective instructions and the right labeling interface is also determinant for success~\citep{Roh:2021}.

Then, {active learning} focuses on iteratively selecting the most informative (according to some pre-defined measure) unlabeled examples for the model to reduce the need for human labor, which can then be outsourced or crowd-sourced~\citep{Roh:2021}. The workers are expected to be accurate, thus the key challenge is to choose the right examples given a limited budget. In addition, semi-supervised learning can complement active learning~\citep{Jin:2014,Camargo:2020} by finding the predictions with the highest confidence and adding them to the labeled examples (as pseudo-labels), while active learning can identify the predictions with the lowest confidence and send them for manual labeling.

Finally, {weakly supervised learning}~\citep{zhou2018brief,Zhang:PAMI:2021} can also reduce the amount of human labor to annotate the training samples. This approach is especially useful when there are large amounts of data, and manual labeling becomes infeasible~\citep{Roh:2021}.
In weakly supervised learning, it is possible to automate the labeling process by defining a set of {labeling functions}~\citep{ratner2017snorkel}, hand-crafted rule-based classifiers. An example from the Snorkel tutorial\footnote{\href{https://www.snorkel.org/use-cases/01-spam-tutorial}{https://www.snorkel.org/use-cases/01-spam-tutorial}} is a labeling function to sort emails as ``spam'', using simply the presence of ``http'' in text meta-data. A labeling function can leverage meta-data or classifiers trained previously on similar tasks. Using multiple labeling functions helps to obtain a label score, which can be interpreted as a label probability, serving as weak supervision.

In conclusion, data annotation can be a time-consuming and expensive process, with many challenges~\citep{s22041596}. However, quantity in some types of machine learning such as deep neural networks is a key to success. Therefore, a lot of research is trying to alleviate this limitation by learning representations from unlabeled data which is later discussed in Section~\ref{chap3:sec:datarepresentation}.

\subsection{Data Transformation} \label{chap3:sec:datatransformation}

Once the data collection process has provided a set of initial raw data, different transformation techniques can be used to make the data suitable for a particular machine-learning model. This section presents a brief overview and discussion around distinct aspects of data transformation, which includes data {integration} or {fusion}, {cleaning}, {reduction}, {representation}, {normalization} or {calibration}, and {augmentation}.

\subsubsection{Integration and Fusion}
\label{chap3:sec:dataintegration}
Data integration or data fusion refers to the process of merging data or datasets from various sources into one dataset~\citep{bleiholder2009data}. For instance, if data are from the same relational database (\eg SQL) but from different tables, then the JOIN operation\footnote{\href{https://en.wikipedia.org/wiki/Join_(SQL)}{https://en.wikipedia.org/wiki/Join\_(SQL)}} is a way to perform data fusion (which can expand sample and feature dimensions). In this case there often exists some matching identifier that helps perform this task directly. Moreover, it is now possible to collect data from external databases (without matching identifiers), websites, or search engines in order to improve the predictive performance of machine learning models.

The type of algorithm used to perform data integration varies depending on the data type (\eg tabular, image, sequence). During such operation, some typical problems to resolve are identification of matching entities (\eg tabular), re-scaling (Section~\ref{chap3:sec:datareduction}), spatial/temporal alignment or registration (\eg 3D shapes, images, videos, Section~\ref{chap3:sec:datanormalization}), cleaning (\eg removal of duplicates, imputation of missing values, Section~\ref{chap3:sec:datacleaning}), calibration and normalization (Section~\ref{chap3:sec:datanormalization}).

This is why machine learning algorithms~\citep{survey_datafusion_ml} can be used to help perform data integration or fusion. For example, the problem of 3D shape registration is usually resolved through the iterative closest point algorithm~\citep{4767965} which is based on the method of least-squares. An advantage of the iterative closest point algorithm is that it iteratively estimates matching points unlike the Procrustes-based method~\citep{10.1093/acprof:oso/9780198510581.001.0001}. Another example is the Fuzzy-Join~\citep{5767865} literature which intends to resolve the problem of inexact matching to merge two different datasets of tabular data.

It is important to mention that despite all the efforts, data integration is often far from perfect and one should keep track of the source of each data as part of meta-data to identify possible biases (\eg the situation where all recorded patients with a given disease are coming from the same hospital).

\subsubsection{Cleaning}
\label{chap3:sec:datacleaning}
Data cleaning refers to the process of improving the consistency of data~\citep{ridzuan2019review}. Some data records may be corrupted (\eg download errors), incoherent (\eg the same entity is represented by different tokens, or the same data has different associated labels), and may include missing data (\eg partial information about a user). Data cleaning can be applied both on the input data (\eg images, videos, text, etc.), the meta-data, or any annotation (\eg the target variable), jointly or separately. Processing all simultaneously is necessary to detect sampling bias with respect to individual samples or groups of samples and decide if the chosen data cleaning method is appropriate.

In fact, distinct methods have been proposed in the literature to deal with missing data such as partial deletion, statistical imputation, interpolation, and Bayesian inference. However, missing data is problematic due to the risk of bias, which depends on the type of missing values, the relative size of the data that are missing, and the way of dealing with these missing values which can have associated risks (\eg yield false positives) and benefits (\eg reduce false negatives)~\citep{SEIJOPARDO201997,SeijoPardo2018AnalysisOI}. For example, if missing data are missing at random then they can be imputed~\citep{van2018flexible}. But, if they are not missing at random, spectrum bias may be reinforced (\eg increase bias toward already present patterns) with imputations. Similarly, samples with missing data could be removed~\citep{guyon1996discovering} but with the risk of introducing an exclusion bias. 

Among data cleaning methods, Bayesian inference~\citep{pmlr-v130-lew21a} is a family of methods offering good performance and automation capabilities. It can also leverage prior knowledge coming from domain expertise (\ie understanding of the data) which is often a key to success.

\subsubsection{Reduction}
\label{chap3:sec:datareduction}

Data reduction corresponds to processes that reduce the information contained in data. Data reduction includes methods to reduce dimensionality in feature space (often referred to as dimensionality reduction), in the number of samples (often referred to as sub-sampling). In some cases, data reduction can also refer to the re-scaling or re-sampling of signals according to spatial or temporal dimensions.
Contrary to data augmentation and feature engineering, the goal of data reduction is to {\em reduce} spurious information in data, for instance, to accelerate learning, or to make the predictor more robust by eliminating redundancy or noise in data.
While several machine learning algorithms are naturally designed to separate important patterns from noise in the data, this can be further aided by data-centric approaches such as data reduction. 

With respect to features, a canonical way to perform dimensionality reduction is the Principal Components Analysis~\citep{wold1987principal} (PCA) which consists of building a subset of features, which are linear combinations of the original features and explains best the variance in data. PCA can be viewed as an ancestor of neural network auto-encoder methods for manifold learning. Indeed, \citet{bourlard1988auto} have shown that for $n$ inputs, a 2-layer network with $d<n$ hidden units, trained to reproduce its input on its output with mean-square-error, yields a representation projecting the data in the $d$ directions of largest variance. Non-linear auto-encoders and their descendants (such as denoising auto-encoders~\citep{NIPS2013_559cb990} and variational auto-encoders~\citep{kingma2013auto}) provide a generalization of PCA. However, sometimes when working in a reduced space it is then required to reconstruct data in the original space. This step can be challenging with autoencoders and in many cases, reconstructed data do not satisfy {validity constraints} (\ie data are not realistic). Some methods such as grammar autoencoder~\citep{pmlr-v70-kusner17a} impose additional structure to alleviate this challenge.

Feature selection methods~\citep{guyon2003introduction} are a special case of dimensionality reduction methods that avoid replacing features with newly constructed features, which can facilitate explainability (\ie interpretation about how a prediction is constructed from the inputs) in some applications. Like other methods of dimensionality reduction, removing redundancy and noise is the primary goal.

Also, more classical quantization can be performed to compress a signal. Quantization is the process of mapping a variable from an uncountable (\eg real values) to a countable space. It is the core of lossy-compression algorithms and can be used to reduce the memory size of signals. A typical and fast way to perform quantization is through the $k$-means algorithm~\citep{pollard1982quantization}.

Some dimensionality reduction methods are directly applicable to reduce the number of samples such as PCA or clustering methods~\citep{yen2009cluster}. It is also possible to leverage gradient-based methods to detect non-informative samples~\citep{killamsetty2021grad}. While sub-sampling can be used to balance the proportion of different classes in a dataset, it is not clear if it is consistently well-performing~\citep{garcia2012effectiveness}. The Imbalance-Learn Python package~\citep{JMLR:v18:16-365} provides a set of algorithms to perform sub-sampling.

The utility of data reduction is particularly important for spatiotemporal data which quickly grow in size and face computational and memory limitations~\citep{10.1145/3439334}. Although many of the previously introduced areas of data reduction have been extended to this setting, down-sampling of spatial resolution with bi-linear interpolation and strided sub-sampling over the temporal axis are often preferred in practice.

\subsubsection{Representation}
\label{chap3:sec:datarepresentation}

Data representation refers to a set of techniques that attributes a numerical representation adapted to learning and more specifically a model. Basic data types can be real (\eg height of a person in centimeters), discrete (\eg number of users on a website), categorical nominal where there is no order on categories (\eg type of vehicles such as car, scooter, or truck) and categorical ordinal where there is an order on categories without a clearly defined numerical scale (\eg rating of a restaurant such as `very bad', `bad', `medium', `good' or `very good'). In machine learning, data are generally represented as tensors (\ie a $n$-dimensional matrix) even in the case of non-regular structures such as graphs (\ie node-features, edges-features, connectivity which corresponds to $n=3$). The problem of finding a good representation is key in machine learning (\ie a representation that makes the model learn and generalize better).

In some cases, one wants to convert images into more high-level features. 
While techniques like deep neural networks can indeed learn directly from the raw pixel values, there might be reasons to prefer more abstract and, semantically richer, higher-level representations.
A straightforward way of obtaining those is to use pre-trained models~\citep{weiss2016survey} (benefiting from transfer-learning), such as I3D~\citep{Carreira_2017_CVPR} or R(2+1)D~\citep{Tran_2018_CVPR} for spatio-temporal feature representations. By pooling the representation provided by the penultimate layer of a deep neural network trained on a source task, one can append to it a new layer to be trained for another target task, such as by adding a classification layer (\ie linear mapping with a Softmax function). This strategy, namely {linear probing}, provides an alternative to training from scratch new models or even avoiding {fine-tuning} (\ie where all weights are trained for the target task) those same pre-trained models leveraged for feature extraction. In particular, for many computer vision tasks, such a strategy provides a good enough initialization and speeds up the training on new datasets, while being perhaps the most convenient strategy for small vision datasets.

Then, to overcome the difficulties of data annotation, by reducing the quantity of annotated data, a lot of research has focused on learning representations from unlabeled data. Learning representations from unlabeled data is now called self-supervised learning~\citep{ZHAO2024122807} but directly corresponds to an extension of works previously classified under unsupervised learning. Self-supervised learning had many successes in natural language processing with methods such as Word2Vec~\citep{mikolov2013efficient}, where a vector representation of words is learned, and arithmetic can be performed on such vectors having some plausible semantic interpretation. For instance, subtracting the vector representing ``men'' from that representing ``king'', then adding those of ``women'', yields a position in vector space close to ``queen''. 
The Word2Vec representation is obtained with a neural network, having at their input a part of the sentence, except for a missing central word, and at the output the central word to be predicted (this algorithm is referred to as Continuous Bag of Words). 
Word2Vec has been a leap forward compared to previous bag-of-word representations, only based on frequencies of words in documents, such as TF-IDF~\citep{tfidf}. Many other works followed the steps of Word2Vec and brought new achievements in the area of natural language processing (\eg Glove~\citep{pennington2014glove}, fastText~\citep{bojanowski2017enriching}, BERT~\citep{devlin2018bert}, RoBERTa~\citep{liu2019roberta}, XLM-R~\citep{conneau2019unsupervised}). More recently, large language models have appeared~\citep{gpt3} as the final realization of this methodology. 

In computer vision, self-supervised learning is also applied to extract representations of images, which are later fine-tuned for supervised classification or regression tasks. In fact, using self-supervised learning on medical images has been shown to reduce the learning of spurious correlations~\citep{goel2020model} between input data and annotations, which also results in better performance (about +10\% more on accuracy in some cases)~\citep{azizi2021big}. The ideas from natural language processing and computer vision are now being generalized to other data representations such as graphs (\eg Graph2Vec~\citep{graph2vec_NarayananCVCLJ17}) and spatiotemporal data.

All primordial methods of self-supervised learning are variants of auto-encoders, which learn a latent representation by first encoding and then decoding. They have recently been renamed ``non-contrastive self-supervised learning methods''. A recent methodology for non-contrastive self-supervised learning is to use in-painting to learn to predict missing parts (\ie occlusions) of an input image, where the occluded image is at the input of an auto-encoder and the missing part(s) at the output~\citep{pathak2016context}. The limitation of non-contrastive self-supervised learning methods is that the model is not informed about {counter examples}, \ie examples which are out-of-distribution or out of the support of the positive examples provided. This motivated the need for contrastive learning self-supervised learning~\citep{chen2020simple}, based on pairwise comparisons of similar and dissimilar examples. To that end, a Siamese neural network architecture~\citep{bromley1993signature} is used, consisting of two identical networks whose outputs are compared with a contrastive loss function, such that agreement is maximized for similar or compatible inputs (\eg two images of the same object, but from different views) and minimized in the case of dissimilarity or incompatibility (\eg inputs represents different objects). 

\begin{figure}[htbp]
    \centering
    \includegraphics[width=0.5\textwidth]{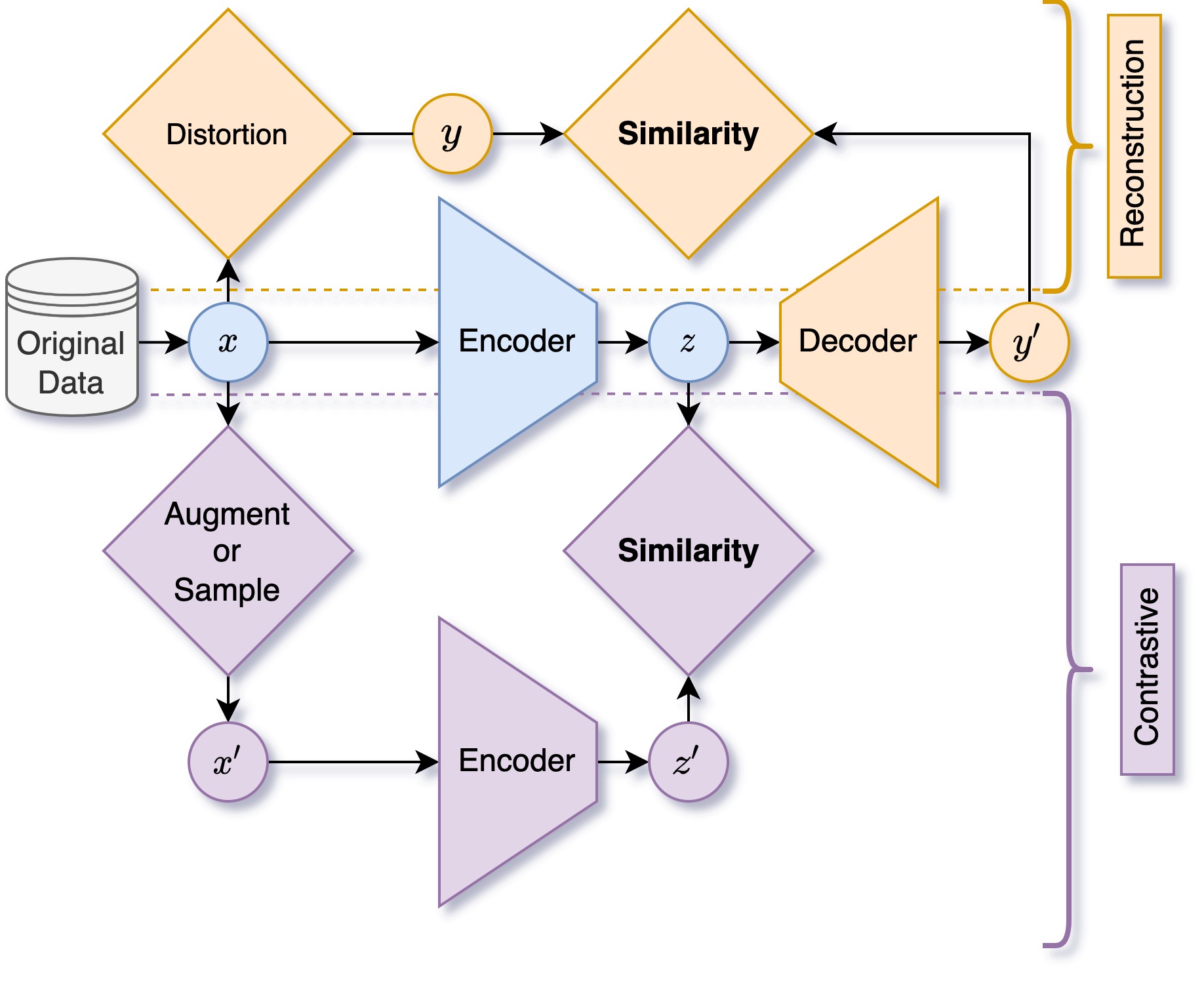}
    \caption{Self-Supervised Learning through Contrastive (purple) or Non-Contrastive (orange) Learning. The input data is $x$ and the representation learned is $z$ for both.}
    \label{fig:self-supervised-learning}
\end{figure}

We illustrate the similarities and differences of self-supervised learning methods in Figure~\ref{fig:self-supervised-learning}. Both have in common the mapping of the input data $x$ to a new representation $z$ (in blue). For non-contrastive (orange), often based on reconstruction schemes, the goal is to learn the representation $z$ from $x$ which helps reconstruct the distorted data $z'$ (\eg jigsaw puzzle, in-painting). For contrastive (purple), the goal is to learn similar representations for similar entities and different representations for different entities. Similar entities are artificially simulated with data augmentation (\eg $x$ can be the image of a bird, and $x'$ is the image of the same bird with a rotation), and then a contrastive loss can be used to enforce the contrastive idea. 

\subsubsection{Normalization and Calibration}
\label{chap3:sec:datanormalization}

Data normalization\footnote{not to be confused with ``database normalization''} or data calibration aims to get rid of some systematic bias, which may occur in data collection, due to a number of uncontrolled factors (\eg a change in operator, a change in temperature, humidity, luminosity, amount of a certain reagent, etc.). Data calibration is not to be confused with model calibration which focuses on calibrating predictions to improve their probabilistic interpretability.

Linear normalizations or calibration simply amount to shifting and scaling features, by an amount determined either by comparing samples to one another (\eg normalization by dividing by the maximum), by the sample itself (\eg normalization by diving by the norm of the sample) or by using some reference values (calibration). For example, standard feature normalization consists of removing the mean and dividing by the standard deviation feature-wise. This is a common pre-processing to neural network inputs because all features are then made unitless and spread over a similar range. In normalization, the quantities used to preprocess the data are directly estimated from these data. Therefore, one needs to be aware that the smaller the dataset is, the higher the variance of these estimates is (\ie standard error is a function of the $\frac{1}{\sqrt{n}}$ where $n$ is the number of examples), this can create unstable results when the dataset is small.
In tabular data, normalizations are often carried out row-wise or column-wise, or both, depending on the nature of the application. Some practical way of performing normalization is through the Scikit-Learn\footnote{Scikit-Learn: \href{https://tinyurl.com/bdfr7z62}{https://tinyurl.com/bdfr7z62}} library which provides ready methods more or less sensitive to outliers such as: MinMaxScaler, MaxAbsScaler, StandardScaler, RobustScaler, Normalizer, QuantileTransformer and PowerTransformer.

Calibration can be thought of as a learning problem, with a very small training set. There are two types of calibrations making use of either internal or external calibrants. An internal calibrant is included in every sample. For instance, in chemistry this would be a compound spiked in known quantity in a sample to adjust the scale of a titrating device; in photography, this would be \eg landmarks positioned with a given geometry or a known pattern of given shapes and colors, captured together with the scene, serving as a reference to compensate for camera aberrations and/or adjust the color spectrum of pictures taken. In contrast, an external calibrant is a reference sample (with a well-defined pattern) inserted regularly in between regular samples. For instance, a chessboard image in photography, or the use of a water-solution in a spectrometer. 

Internal calibrants are used when measurements are constantly changing, while external calibrants are suitable for slow drifts in recording conditions. In either case, the ground truth (target values) for the calibrants are known. This allows users to train a very simple predictive model (often just linear), which maps measured values to target values. The predictor can then interpolate between known values to correct the other measured values in the sample.
This type of method is commonly used in computer vision for camera calibration~\citep{zhang2000flexible}. Calibration is particularly needed for data fusion when samples are obtained from different sources, which relates this problem to data integration introduced in Section~\ref{chap3:sec:dataintegration}. Another example is the calibration of data from multiple sensors such as in autonomous vehicles~\citep{yeong2021sensor}.

\subsubsection{Augmentation}

Data augmentation is the process of artificially increasing the size of an already existing dataset (either with respect to samples or with respect to features~\citep{Liu_2018_CVPR}). It potentially enlarges the dataset by orders of magnitude, at a reduced cost. It can be {rule-based}, such as in computer vision, where it is possible to perform changes in resolution, orientation, brightness, execute random crops/shifts, and include noise, among others~\citep{Shorten2019}. It can also be {learning-based}, where the underlying distribution of the data is learned via generative models (\eg generative adversarial networks~\citep{goodfellow2020generative}, variational auto-encoders~\citep{kingma2016improved}), to then artificially sample from it. Although data augmentation can lead to overall improvements in performance (\ie improved average metrics), it may introduce {\em selection bias} yielding an increase in performance for some classes or groups, at the expense of others. This is related to the fact that designing dataset/task-specific regularizers, without introducing selection bias, remains an open research question~\citep{Balestriero2022}. That is why proper model evaluation and selection needs to be conducted to quantify such effects. Some types of data augmentation are preferably executed during the training to avoid storing the augmented data. Finally, augmentation is often used jointly with other learning methods, especially self-supervised methods such as contrastive learning~\citep{chen2020simple} and siamese networks~\citep{simsiam_chen_2020} to leverage the knowledge of proximity between original and augmented samples.

\section{Evaluation}
\label{chap3:sec:evaluation}

The goal of dataset evaluation is to assess whether a dataset meets its original dataset requirements. 
This includes not only verifying that specified quality and quantity criteria are met but also catching errors in dataset implementation or flaws in the dataset design.
In this section, we propose to evaluate a dataset based on
(i) soundness (which we will identify by looking for inconsistencies, and, detecting and correcting confounding bias), 
(ii) completeness (which we will identify with detecting and correcting selection bias), 
as well as evaluating the protection of human subjects with respect to (iii) fairness and (iv) privacy, if applicable. 
Any of these criteria can benefit from qualitative (\eg through data visualization) and quantitative assessments (\eg through metrics).
\begin{figure}[htbp]
    \centering
    \includegraphics[width=0.8\linewidth]{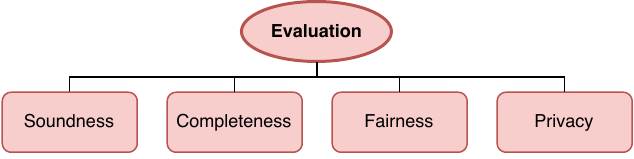}
    \caption{Categorization of sub-processes included in data evaluation.}
    \label{fig:dataset-development-evaluation}
\end{figure}

\subsection{Preliminaries: Inspection, Visualization, and Baselines}
\label{chap3:sec:inspection-visualization-baselines}

\begin{figure}[!ht]
    \centering
    \includegraphics[width=0.32\textwidth]{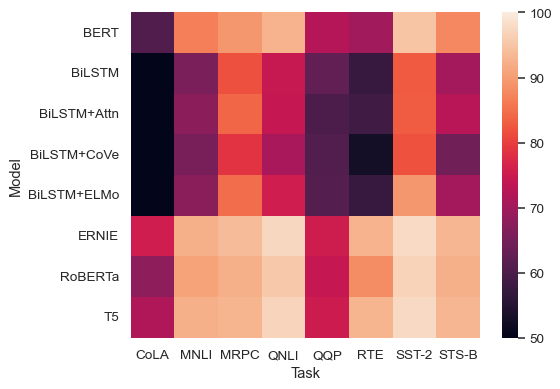}
    \includegraphics[width=0.32\textwidth]{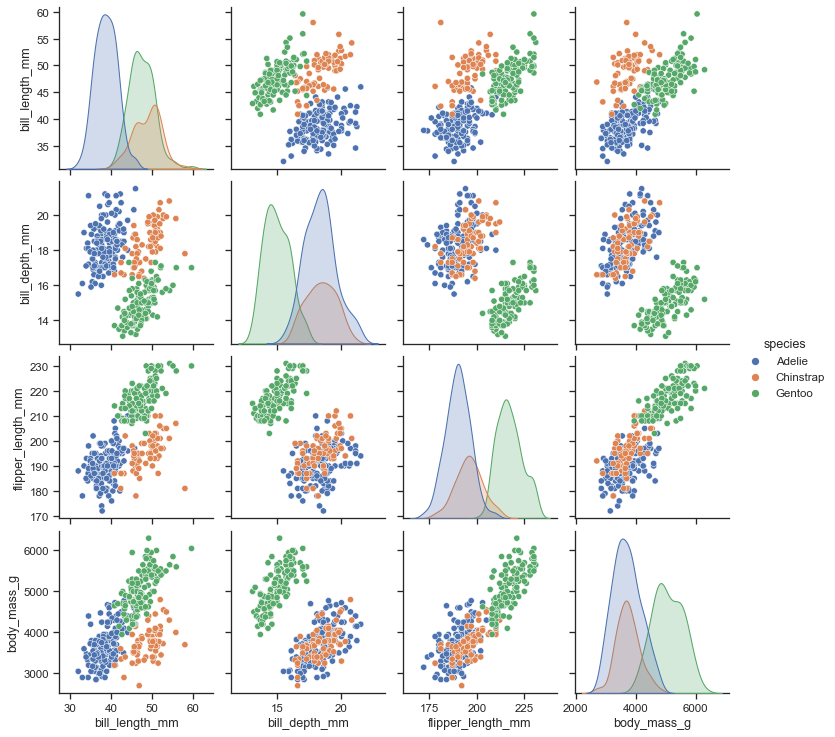}
    \includegraphics[width=0.32\textwidth]{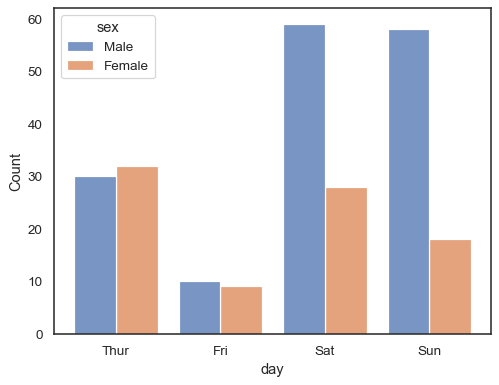}
    \caption{Example visualizations of (left) a heatmap, (middle) a pair-plot, and (right) class frequencies from Seaborn.}
    \label{fig:example-visualization}
\end{figure}

Visualization is key in evaluating a dataset (see Figure~\ref{fig:example-visualization}). It is recommended to prepare visualization tools that your data collectors can use during data collection (Section~\ref{chap3:sec:datacollection}):
\begin{description}
  \item[Contact sheets:] For image or video data, samples must be visualized \eg in contact sheets showing a subset of examples of each class. For non-visual data, find a representation for each sample, which is visual (a spectrogram or an embedding). This will then also be helpful to visualize later misclassified examples.
  \item[Heat maps:] Once the dataset is in vectorial representation, heatmaps or cluster maps are useful to check whether there is no anomalous structure in the data.
  \item[Pair plots:] For datasets with a small number of features, pair plots allow users to visualize whether classes are easily separated by unique features or pairs of features. If the dataset has a large number of features, you may want to first apply PCA and then do pair plots in the first few principal components. Note that this is just a tool for discovering general trends, as complex concepts are rarely separated by a subset of two features.
  \item[Class frequencies:] Use bar graphs to represent \eg the frequency of examples in each class.
\end{description}

Some commonly used library for data visualizations are: Matplotlib\footnote{\href{https://matplotlib.org/}{https://matplotlib.org/}}, Seaborn\footnote{\href{https://seaborn.pydata.org/}{https://seaborn.pydata.org/}}, and/or Microsoft SandDance\footnote{\href{https://microsoft.github.io/SandDance/}{https://microsoft.github.io/SandDance/}}.

It is recommended to involve baseline methods solving the tasks early in the data collection process. 
Baseline methods for checking the dataset can range from very basic to state-of-the-art. 
Analyzing the results indicates how well current methods perform on the data. 
In particular, the performance gap between the constant predictor and state-of-the-art methods will indicate whether there is a learnable concept in the data. 
Beyond this, it is also a good practice to look at confusion matrices for classification tasks. For example, it can help determine unbalanced predictive performance between different classes.

\subsection{Soundness} \label{sec:soundness}

A dataset is considered sound if it correctly results from its premises, which are the requirements, design, and implementation steps, and if these premises are themselves correct. 
For example, a design choice could be to identify cities by name and location (which would be correct), while another design could be to identify cities solely by name (which would be incorrect because names are not unique identifiers). 
Therefore, this will include verifying upstream steps as well as looking for inconsistencies, corruptions, and a good state of collected distributions.
Next, we discuss some sources of consistency that should receive special attention. 

\begin{description}

\item[Representation consistency] relates to the problem of having a unique representation for the same entity across the dataset. For instance, when collecting articles from the press on the Internet the same city can be written differently such as `New York', `N.Y.', and `the Big Apple', which all represent the same city. Similarly, it is important to check whether physical quantities are all measured using the same unit. In the case of tabular data and storage systems, improving representation consistency is often referred to as data deduplication~\citep{xia2016comprehensive}.

\item[Labelling consistency] refers to self-agreement and inter-agreement among annotators, which is especially important when labeling was performed through crowd-sourcing as mentioned in Section~\ref{chap3:sec:dataannotation}. 
If various annotators are participating in the annotation process, it is important to ensure that they agree on the exact concept that they are labeling, and use the same definition and thresholds.
Self-agreement is particularly useful to identify low-quality annotators while inter-agreement is good to estimate the difficulty of the task.
Krippendorff's Alpha-Reliability~\citep{krippendorff2011computing} is an example of such as metric.

\item[Outlier detection] relates to the identification of observations, which appear to be inconsistent with other observations of the dataset~\citep{hodge2004survey}.
Data visualization is particularly useful to detect the presence of such samples. 
After identifying outliers, a domain expert can decide whether they result from errors.

\item[Bias detection] in general, concerns the identification of systematic bias due to the lack of control or randomization of some nuisance factor. 
Collecting appropriate meta-data is an important tool to help detect bias, including potential nuisance variables (temperature, humidity, luminosity, recording time, date, collection operator, etc.) and protected variables (age, gender, ethnicity, etc.) involved in societal bias. 
Subjective bias (discussed in Section~\ref{chap3:sec:fairness}) coming from data annotation can also benefit from meta-data information as it can be used to reveal different types of bias with respect to both the annotator and the data being annotated~\citep{Junior_2021_WACV, Escalante:TAC:2020}.
A machine learning model may then be trained using variables that are a suspected cause of bias only, or using them together with other predictor variables. Feature selection methods are then applied to determine whether any such variables are significantly predictive. 

\begin{usecase}{Exploiting Bias in a Challenge}
    The ChaLearn LAP Challenge on Self-Reported Personality Recognition~\citep{pmlr-v173-palmero22b} adopted a dataset composed of large amounts of data (audio-visual, transcripts, meta-data, etc.). However, one competitor team was able to achieve promising results by analyzing the correlation between meta-data and the self-assessment personality trait scores (the target variables), while proposing to use a random forest regressor trained solely on meta-data features (\ie age, gender, and number of sessions). That is, the competitor did not make use of the more than 60 hours of audio-visual data available and associated transcripts for training, which data creators believed to be crucial to addressing the problem. In other words, the data creators were not expecting that a model trained on meta-data features could accurately predict the personality of someone, indicating that the adopted dataset may include unwanted bias.  According to this competitor (and challenge results), a simple random forest regressor based on meta-data features only should not be enough to outperform a method based on linguistic, audio-visual, and meta-data features like the alternative model they evaluated in such a complex task as personality recognition.
\end{usecase}

\end{description}

\subsection{Completeness}
\label{chap3:sec:completeness}

The completeness of a dataset is the attribute of a dataset to contain all required features describing sufficiently the problem at hand, as well as to properly select its samples (\eg the number of i.i.d - independent and identically distributed - samples directly impacts the estimation of the mean estimate - standard error). Therefore, completeness can be evaluated with respect to samples or features but also with respect to hidden variables, which do not define the problem but have to be uniformly sampled to avoid bias.

\begin{description}

\item[Feature-wise] completeness is associated with the notion of causality which defines that a cause can be {necessary} or {sufficient}. If $x$ is a {sufficient cause} of $y$, then the presence of $x$ necessarily implies the subsequent occurrence of $y$. However, another cause $z$ may alternatively cause $y$. Thus the presence of $y$ does not imply the prior occurrence of $x$. Then, if $x$ is a {necessary cause} of $y$, the presence of $y$ necessarily implies the prior occurrence of $x$. The presence of $x$, however, does not imply that $y$ will occur. These relations matter to understand the problem of confounding factors, which relates to a false association between two variables such as $X$ causing $Y$ because of a third missing variable $Z$ causing the two~\citep{confounding_bias}. 

Confounding can be divided into omitted variable and correlated noise (also denoted as a spurious feature). As an example of omitted variable: if $X$ is ``drinking coffee'' and $Y$ is ``developing a lung cancer'' some data can show that drinking coffee increases the risk factor of developing lung cancer. However, this is happening because a common cause $Z$, which is ``smoking'', was not taken into consideration and is associated with both ``drinking coffee'' and ``developing lung cancer''. An example of correlated noise is the background (\eg road, sky, water) with the type of vehicle (\eg car, plane, boat). A picture of a car will rarely happen with a sky background and therefore the background (the road) is predictive of the class of the object (the car).

It can be tracked by identifying whether some common candidate confounding factors, which are spuriously predictive of the target variable (individually or jointly). For sensor data, typical examples are temperature, humidity, luminosity, etc., and for survey data, typical examples are age, gender, ethnicity, etc. 
Classical feature importance methods can track how significant such associations are \citep{altmann2010permutation}. 
Proper collection of meta-data (also called protected attributes in fairness) can aid in this process~\citep{aif360-oct-2018}.
If some of these attributes cannot be recorded directly, then it is advisable to try to measure them indirectly. 
For example, if one suspects that the image background could be a spurious feature, then one may create mini-images containing only the background of the original images and try to predict the target variable from them (excluding the object of interest). If the model now predicts the object of interest, this means that the background was utilized in the predictive model~\citep{tian2018eliminating}.
If confounding bias is identified, the data collection process must be revised to alleviate it. 
Missing variables are often the cause of excessive aleatoric uncertainty (or intrinsic randomness), which is the variability of the outcome given input variables because of unknown source factors. 
This variability needs to be assessed to make sure the prediction is performed within a reasonable confidence interval.

\item[Sample-wise] completeness is directly associated with the problem of selection bias, which can be divided into exclusion bias (removal of samples) and spectrum bias (only a subset of the target population was observed). Exclusion bias comes from a choice of the dataset development team. 
For example, it can be a result of data cleaning, which may remove too many samples or by removing them (possibly for good reason) may modify the data distribution, \eg create an imbalanced dataset. 
It can also be a choice of filter in a search engine or a selected window of observation. 
A well-known bias in an academic dataset is indeed the recruitment of graduate students as subjects. Testing sample-wise completeness depends on the assumption made on the problem. In the i.i.d. samples case it can be tested through the classic generalization error, in the context of combinatorial (or compositional) generalization the test set would require to contain unobserved combinations with already observed features. Similarly, the test will vary for spatio-temporal data and possible distribution shifts.
The question of whether more data can further improve the learning algorithms is often addressed by using learning curves of machine learning algorithms~\citep{mohr2022learning,mohr2023fast}.

\end{description}

\subsection{Fairness}
\label{chap3:sec:fairness}

Fairness has recently attracted attention in the machine learning community~\citep{Bird:2019:FML,Mehrabi:2021} after several flaws arising from misuse of biased data being reported by the media such as the Guardian, 2022\footnote{The Guardian, 2022: \href{http://tinyurl.com/yfeeu2sx}{http://tinyurl.com/yfeeu2sx}}  or in the Washington Post\footnote{The Washington Post, 2020: \href{http://tinyurl.com/4kv4nf5v}{http://tinyurl.com/4kv4nf5v}}.
In the context of decision-making, and according to ~\citep{Mehrabi:2021}, fairness is the absence of any prejudice or favoritism toward an individual or group based on their inherent or acquired characteristics.
Therefore, an unfair algorithm is one whose decisions are skewed toward a particular group of people. 

Machine learning algorithms can inherently transfer bias from data to model.
When black-box models are used (\eg most deep learning architectures), results are usually difficult to explain and interpret, making bias mitigation even harder. Hence, a biased dataset used for training or evaluating machine learning methods can negatively impact outcomes. For instance, a face recognition method trained on male faces will probably not generalize well to female faces. To mitigate such problems, in most countries, the law protects against a discriminatory decision (\eg about hiring or condemning), based on {protected attributes}, which include gender, age, and ethnicity, among others~\citep{barocas-hardt-narayanan}. 
One way of ensuring some level of fairness at the data collection level is to ensure that samples are group-balanced, \ie that there is an approximately equal number of samples in all groups resulting from combinations of protected attributes and labels.
Of course, this may be impractical, and it may be more feasible to record the protected attribute to later correct bias at the machine learning level, with in-processing~\citep{10.1145/3551390} (\ie, during learning, which is categorized into explicit and implicit, where the former directly incorporates fairness metrics in training objectives, and the latter focuses on refining latent representation learning) or post-processing techniques (\ie after learning). However, recording protected variables may raise issues of privacy (Section~\ref{chap3:sec:privacy}).

Data annotation may also be a source of social bias.
The machine learning and computer vision communities are starting to pay more attention to this problem, as it relates to fairness~\citep{Bird:2019:FML}.
Recent work reports different types of subjective biases coming from crowd-sourced annotations.
Although biases produced by human perception have been widely studied in sociology and psychology (\eg gender~\citep{todorov:2019:gender} or attractiveness~\citep{Talamas:2016} bias), little attention has been given to subjective bias analysis~\citep{Shen:2019:ACII,Quadrianto_2019_CVPR,Robinson_2020_CVPR_Workshops,Yan:ICMI:2020} beyond the perspective of explainable models~\citep{Escalante:IJCNN:2017,Park:CVPR:2018,dario:TAC:2019,Escalante:TAC:2020}. Moreover, as perception is dependent on the observer, the relationship between annotators and the entity being annotated could explain how some perception biases are produced, which is an almost unexplored area in computer vision and machine learning. However, this would require a dedicated discussion around privacy and ethical issues~\citep{Junior_2021_WACV}.

Over the past few years, a vast number of scientific events and studies appeared intending to stimulate discussion and advance the state of the art on fairness and bias mitigation methods (\eg ACM FAccT\footnote{\href{https://facctconference.org}{https://facctconference.org}}), explainability and interpretability (\eg \citet{Escalante:IJCNN:2017,Park:CVPR:2018}).
Distinct definitions and metrics for fairness have been proposed and discussed, like ``fairness through unawareness'', ``individual fairness'', ``demographic parity'', ``equalized odds'',  ``equality of opportunity'' or ``counterfactual fairness''~\citep{NIPS2017_a486cd07,ASHOKAN2021102646,aif360-oct-2018}.
Although there is no standard definition of fairness that could be used for all types of problems, researchers must be attentive to possible fairness issues and consult with social scientists and ethics specialists, as needed.
For instance, it is advisable to have data collection protocols reviewed by an Institutional Review Board (see Section~\ref{chap3:sec:requirementsanalysis}) even though this does not guarantee to solve all possible problems.

\subsection{Privacy}
\label{chap3:sec:privacy}

When a dataset contains human-related data, privacy, and data protection become mandatory and will impact all aspects of the dataset development.
Data protection regulations (\eg the European General Data Protection Regulation, GDPR~\citep{10.5555/3152676}) define different levels of protection depending on each type of data, and each level has different requirements for processing and storing.
Moreover, the classification of personal data is different in each regulation and is something that can change regularly.
The inclusion of such human-related data will also require collecting an explicit consent form before capturing any data, with a clear and understandable description of the collected data, how it will be used, who will have access to it, and for how long the data will be stored.

Data are considered anonymized if there is no possibility to identify a subject using the provided data.
In cases where data curators maintain a correspondence between published data and originally captured data, we cannot consider the data anonymized, as there exists the possibility to recover the identity of a subject.
In this case, the data is pseudo-anonymized, where without the link between real identity and published data, no one can recover the real identity of a subject in the dataset.
Pseudo-anonymized data allow data curators to remove the data from a certain individual if it is required but have a security risk since if this link is compromised, real identities can be recovered.

In the context of anonymization and pseudo-anonymization, $k$-anonymity is an important measure.
The measure of $k$-anonymity implies that given one entry of the datasets (\ie information of a certain individual), there exist at least $k-1$ identical entries in the dataset corresponding to other individuals. 
The minimum recommended value for $k$ is three, although bigger values ensure better anonymization. 
For many public datasets, re-identification is easy even when data seems anonymous.
For example, \citet{Sweeney2000} demonstrates that $86\%$ of the U.S. population can be uniquely identified with just three ``quasi-identifiers'' which are the zip code, the gender, and the date of birth.
Although there are easy ways to increase the $k$-value, \eg by binning variables, if a dataset contains sensitive data, it is a good idea to apply one of the many existing anonymization algorithms~\citep{DBLP:conf/mdai/Casas-RomaHT12}.
Finally, it is important to make sure that the anonymization process does not affect the usefulness of the dataset.
\citet{DBLP:conf/cbms/CarmonaCC19} provide an introduction to the subject in the case of health data.

Differential privacy (DP)~\citep{10.1007/978-3-540-79228-4_1} and the possibility of replacing real data with realistic synthetic data providing some guarantees of privacy have gained some recent attention. 
\citet{radioactive_sablayrolles_2020} propose an apparatus in which an ideal attacker having maximum information evaluates whether such synthetic data are protected against membership inference attacks (\ie determining whether a sample was or not part of the data used to train the generative model). 

\section{Distribution and Maintenance} 
\label{chap3:sec:distributionandmaintenance}

Dataset {distribution} is about making the developed dataset accessible to others, while {maintenance} are all tasks and processes required to maintain the dataset accessibility and to perform changes on the dataset or the way it is distributed in order to improve its accessibility or quality. Those two tasks should often be considered together. Depending on the structure of the dataset (\eg size, data type, format) those can be simple or complex tasks. Taking into account the requirements of the dataset (Section~\ref{chap3:sec:requirementsanalysis}), one may select the maintenance and distribution technologies/strategies that are cost-effective, sustainable, and supportable with available resources (economic and human). Next, we present the aspects we consider the most relevant to be considered:

\begin{description}

\item[Ownership and licensing:] The distribution of a dataset includes important legal requirements. The dataset creators have to define the usage and responsibility of distributed data. 
To this end, the dataset can be distributed under copyrights, specific licensing~\citep{Benjamin2019TowardsSO} (\eg open-source\footnote{\href{https://opensource.org/licenses}{https://opensource.org/licenses}}, creative commons\footnote{\href{https://creativecommons.org/}{https://creativecommons.org/}}) or terms of use (ToU). 
Licensing not only includes the dataset creators but also the object of shared data such as information about individuals, whether the information is personal or medical, 
whether the individual agreed to distribute this data or whether regulation exists for this type of data (\eg General Data Protection Regulation\footnote{\href{https://gdpr.eu/what-is-gdpr}{https://gdpr.eu/what-is-gdpr}} in Europe, California Privacy Rights Act\footnote{California Privacy Rights Act: \href{https://tinyurl.com/mr38ctvu}{https://tinyurl.com/mr38ctvu}}). 
Therefore, the maintenance plan may include all the processes required by the data protection regulations, such as the capacity to remove all the data for an individual in case it is required, or the partial or total elimination of the dataset and/or original data. The distribution tools may also support such actions. 

\item[Hosting platforms:] The dataset can be hosted in a Cloud service (\eg Microsoft Azure, Google Cloud, or Amazon Web Services) and benefit from optimized downloading services if the source (where the dataset is stored) and the target (where the dataset is downloaded) platforms are from the same service (\eg Google Drive and Google Colab with \texttt{gdown}). It is also possible to directly store the data on a web server so that the download can be performed through \texttt{http}. Other possibilities are through File Transfer Protocol (FTP) and Secure Copy Protocol (SCP). More recent open-science services such as Globus\footnote{\href{https://www.globus.org/}{https://www.globus.org/}} can also be set up to optimize the transfer of data.

\item[Evolution and versioning:] Datasets often evolve in time, both to fix errors or to include new data or labels. Also, flaws inside the dataset can be discovered later through active usage~\citep{REVISE_imagenet}. A good example is linked to the evolution of machine learning research. Recently the importance of privacy preservation and fairness in machine learning has become a priority. Therefore, the ImageNet dataset was updated to anonymize individuals appearing in pictures or filter out problematic samples\footnote{ImageNet: \href{https://tinyurl.com/54ux9bsu}{https://tinyurl.com/54ux9bsu}}. Users were informed about these changes through the website of ImageNet. 

Due to these continuous changes, it is important to attribute a version or unique identifier to the dataset to be able to differentiate it in case of modification. Open-science (free of access) websites such as Zenodo\footnote{\href{https://zenodo.org/}{https://zenodo.org/}} and arXiv\footnote{\href{https://arxiv.org/}{https://arxiv.org/}} now provide digital object identifiers (DOI) which can be used for this purpose. Sometimes it is also possible to use data version control\footnote{\href{https://dvc.org}{https://dvc.org}} if the data is not required to be completely removed (\eg for scientific reproducibility) so that versions of the data can be tracked. Data version control is essential when having an evolving dataset to keep track of the versions and give the possibility to trace which dataset was used to train a particular model. 

\item[Data format:] When distributing data it is often important to compile the data under a compressed format so that it can be downloaded faster. Such format can be \texttt{.tar}, \texttt{.zip}, and \texttt{.gz} to cite a few. When using a compressed format it is important to inform users about the decompressed size. Although the format can change due to the evolution of the dataset, it is desirable to maintain backward compatibility as much as possible. This is a general software principle also valid for datasets. A big format change can limit its usability. 

\item[Dissemination:] Lastly, in the case of a public dataset it is important to communicate its existence. The organization of related competitions and events in international conferences can be a good opportunity to present the dataset and a first benchmark using it. The NeurIPS conference promoted a specialized track related to the creation of new datasets and benchmarks that can help showcase such datasets. More details are provided by \citet{richard2024ai}.

\end{description}

\section{Discussion}

Datasets are an important aspect of scientific benchmarks and competitions for machine learning. 
More importantly, properly designed and evaluated datasets are of extreme importance for developing trustworthy and robust artificial intelligence systems. 
In this chapter, we have made an attempt at specifying the dataset development process. 
We have categorized the various steps that should be undertaken in the dataset development, \ie, documentation, requirements, design, implementation, evaluation, and, distribution and maintenance. 
We approached dataset development as an agile process that is often iterative and requires interactions between its sub-processes.
However, we acknowledge that every dataset development process can be different, and offers the opportunity to emphasize or skip certain parts of this process. 
For example, in some cases, emphasis is placed on the design phase, whereas in other cases the maintenance phase is limited (\eg when there is no ability to improve the dataset after it has been released). 

While we have attempted to give a broad overview of the dataset development process, this is by no means exhaustive. 
When developing a dataset one has to take certain care to not introduce any type of bias. 
Every dataset development process can introduce its distinct type of bias.
Furthermore, while this chapter focuses on the dataset development process, many machine learning benchmarks and competitions also include the evaluation of models trained on these data. 
Typically, this involves splitting the dataset into a train and test set, which, when not addressed properly, can induce other types of bias (\eg information bias and data leakage). 
This is out of the scope of this chapter.

With this chapter, we aimed to harmonize some terminologies from the dataset development process as well as bring together several directions of the literature that we expect to be taken into consideration when developing new datasets. 

\section{Acknowledgment}

This material is based upon work partially supported by the U.S.\ Department of Energy (DOE), Office of Science, Office of Advanced Scientific Computing Research, under Contract DE-AC02-06CH11357. 
This material is based upon work partially supported by the ANR Chair of Artificial Intelligence HUMANIA ANR-19-CHIA-0022 and TAILOR EU Horizon 2020 grant 952215. 
This work has been partially supported by the Spanish projects PID2022-136436NB-I00 and PID2022-138721NB-I00, and by ICREA under the ICREA Academia program.

\bibliography{ref}

\end{document}